# Exploring the pattern of Emotion in children with ASD as an early biomarker through Recurring-Convolution Neural Network (R-CNN)

Abirami S.P.[a,*], Kousalya G[b] and Karthick R[c]

[a] *Assistant Professor, Department of Computer Science and Engineering, Coimbatore Institute of Technology, Coimbatore, India.*
[b] Professor, *Department of Computer Science and Engineering, Coimbatore Institute of Technology, Coimbatore, India.*
[c] *Occupational Therapist, Steps Rehabilitation Center (I), Steps groups, Coimbatore, India.*

*Corresponding author: Abirami S.P. ; email:* abirami.sp@cit.edu.in

**Abstract**

Autism Spectrum Disorder (ASD) is found to be a major concern among various occupational therapists. The foremost challenge of this neurodevelopmental disorder lies in the fact of analyzing and exploring various symptoms of the children at their early stage of development. Such early identification could prop up the therapists and clinicians to provide proper assistive support to make the children lead an independent life. Facial expressions and emotions perceived by the children could contribute to such early intervention of autism. In this regard, the paper implements in identifying basic facial expression and exploring their emotions upon a time variant factor. The emotions are analyzed by incorporating the facial expression identified through CNN using 68 landmark points plotted on the frontal face with a prediction network formed by RNN known as RCNN-FER system. The paper adopts R-CNN to take the advantage of increased accuracy and performance with decreased time complexity in predicting emotion as a textual network analysis. The papers proves better accuracy in identifying the emotion in autistic children when compared over simple machine learning models built for such identifications contributing to autistic society.

**Introduction**

In India, the rise of autism rate is witnessed over years as specified by Autism society of India [3]. Though the reasons behind this autism seemed to be a still on-going exploration, the identification along with basic biomarker and proper guidance are the target of many researchers in India. Also, in India, there is no specific social, community or population based studies which may provide exact predominance of autism spectrum. Most of the individual in our country with ASD are still not diagnosed properly and majority live a life of ignorance, deprivation and non acceptance nature of the disability [8].

Although the root cause and mechanism contributing to ASD could not be completely identified, it is apparent that the abnormal neurotransmission in brain regions hinder the child from normal behavior and motor actions to an altered behavior [3,10]. This impairment of mental difficulties could be identified with regular periodic analysis of the child communication behavior, social interaction, object identification, emotional sequence, linguistic ability and knowledge adaptation rate [18]. These abilities of the child are observed between the ages of 10 months to 3 years. Owing to the special consideration and sometimes to an extent of negligence, the children fail in thinking them beyond their capability which leads to psychological barriers that is expressed as emotions in them.

The ultimate responsibility in training the autistic children lies with the high priority scale of early identification and exploring its characteristics. Additionally, analyzing the functional complexity such as low and high functioning autism and on least priority lies the appropriate coaching [9]. The clinical procedure has strongly proved that the children respond to the interaction based on their neurodevelopment over ages. The clinical evaluation and analysis is through the regular dyads manner which is considered to be a time consuming process. These clinical evaluations take place with the input from the parents and caretakers [8]. Also, many other identifications methods were adopted to evaluate the nature of autism in every child like vocal analysis, identification of objects, interest factor analysis, etc [1].

With the advent of facial expression contributing towards the publishing of neural changes and responses to an event, the proposed paper builds the system for early identification of autism through expression and emotional analysis. The FER system analyses the various expressions perceived by the children and a comparison of varied expression possessed would result in gaining insight on the pattern of expression and interest in every child. This active learning of the neural network models could be adopted for autism evaluation at a very early stage where the child tends to make expressions towards the mother and the family members between the ages of 4 to 8 months. This holds a pre screening technique to the clinical evaluation that hints the autistic characteristics and does not utterly confine into autism spectrum.

**Summary:** The research paper deals in briefing the introduction and need for the proposal in session 1; the state of art made towards the paper in session 2; the proposed methodology for emotion identification in session 3; followed by the result discussions and conclusion along with future scope in the consecutive sessions.

**State of Art**

Autism being one of the notable neurodevelopmental disorder, the basic reasons, symptoms and combination of varied characteristics of the spectrum is still an ongoing investigation. Though there are various methodologies evolved in exploring autism in children and adults, there has always a high demand for early intervention.[18] This is owing to the rapid increase in the development of intelligence systems. The basic characteristics are evaluated through defined and authorized checklists by clinical experts. This is further extended by many researchers such as incorporating facial expression, voice modulation, automatic movement tracking, brain image analysis and gene expression analysis along with the basic evaluation. Those evaluations have proved that high functioning of autism spectrum disorder will have significantly impaired communication abilities in social environment [3,9].

Among the various disabilities, expression and emotional discharge in high functioning autism depicts a clear picture of the neurodevelopmental state of autistic people. Such exposure of expressions can either be in a contactless environment or can be under the influence of the object they face. It is this important to differentiate between the emotions possessed and the cause of emotion in the autistic people [2].

The children with autism will show varied expressions based on the event and routine that they experience, Van Eylen et al. (2018) [1]. These abilities are difficult to perceive and their ambiguous expression compared to the regular developing TD counterparts to a greater extent as discussed by Guha et al. (2016). The author employs a new computational intelligence method that analyses facial expression to support autism [13]. Though the characteristics and the diagnostic probability of autism varying between genders are based on sensory symptoms, the basic characteristics of autism remain the same at the early developmental stage as discussed by Duvekot et al. (2017) [10].

Classification of facial expressions of the children was focused when the target object possesses anger and happy emotion, Whitaker et al. (2017). Exploring such emotions through facial detections using machine learning algorithms could result in an early identification of autism before the clinical analysis [20]. Though, to further improve the accuracy in classification and better reliability of the screening mechanism, a deeper feature identification and feature analysis should be involved as discussed by the author.

Also, the facial feature specification was proposed and upgraded by researchers that categorized the face into T shaped structure extracting eyes, nose and position in three feature dimensions, Wang &Adolphs (2017). The insights were analyzed based on the facial expression evaluated over the center axis of the face including eyes, nose and mouth. The major difficulty faced was the position of the face with respect in viewing the entire features of the face [19].

Similarly, method that analyzes the geometric facial features to establish the face models of every individual was proposed, Gopalan et al. (2018). This geometric analysis was made with light illumination, orientation and color scaling of the captured image [11]. This did not prove its best to the real time applications, as the considered parameters are harder to achieve in real time applications with varied light rays enlightenment.

The facial expression analysis was experimented with facial region detection using Haar technique that covered the regions of face like mouth and eyes, Bone et al. (2016). The face image was divided into two regions on application of Sobel edge detection that classified the regions with facial features [8]. Similarly various other researchers have given reasonable accuracy for FER systems through techniques such as HOG, SVM, random forest, basic neural network, etc [12,14].

Considering the advantage of deep neural network, attention convolutional network that works in identifying the facial expression, by classifying the regions of face that contribute major towards the expression identification, Shervin at al. (2019) [16]. The basic CNN architecture was improved for its accuracy namely incremental boosting CNN, identity aware CNN, CNN with varied hidden layers thus resulting in modified CNN structure for FER systems.

Likewise, there are various possibilities in inculcating advanced features selection and feature analysis for face detection in practical real time environment. The ultimate aim of the technical improvement is to produce a better solution until the false positive detections are rejected quickly during the early stages, Jain et al. (2018) and Jan et al. (2019) [14,15].

The proposed paper importantly bridges the gap identified in early exploration of autism behaviour as the developmental ability is so far measured using a biomarker on how accurate the autistic children identify the expressions of the human they face. This is to analyze the perseverance and processing capability of the child brain to differentiate between expressions. This kind of evaluation was extended in the view of understanding objects by the children under static and dynamic considerations rather failed to measure the expression and emotion perceived by the children [4,5]. In view of the expression identification, emotion and the social behaviour of the autistic child, the paper aims in building a model for such neural connectivity and articulating nature.

**Proposed Methodology**

With the dawn of the improvisation in the technological field relating to healthcare development in particular to autism, the proposed methodology aims in identifying facial expression and emotion along the time variant. The block diagram for the proposed architecture is shown in Figure 1.

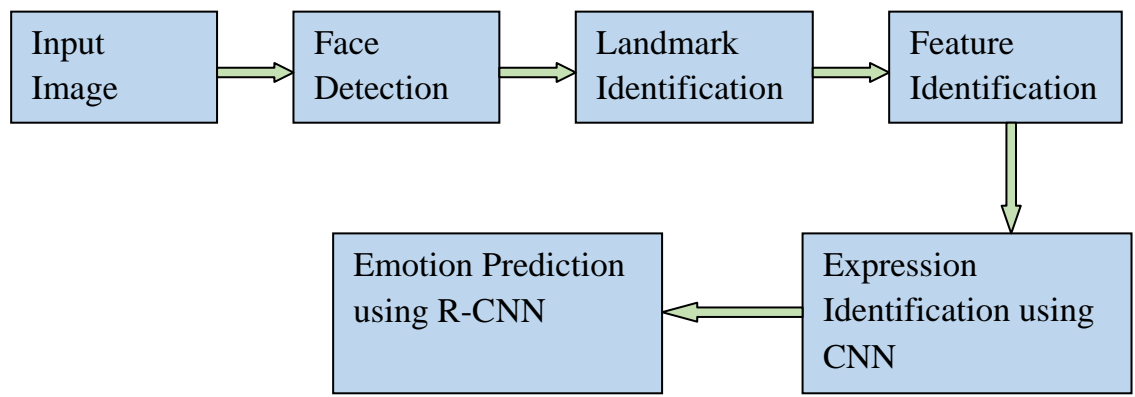

**Figure 1. Basic Architecture of the Proposed RCNN-FER system**

**FER system using CNN**

The facial expression of the children is identified on application of CNN. In reference to our previous work [7], in the proposed technique, the input images were pre processed in resizing it to a 64x64 dimension image. The images were grouped based on the various classes of expressions in their labelled order. The input image is then processed to identify the 68 facial landmark points distributed over the frontal face. The nose points are kept static to analyse the angle of facial position. The neural network is then trained in an order under each category of expressions to identify the important features in the facial structure. The convolution neural network is trained for both ASD positive and ASD negative to check for the training correctness.

The first convolution layer of the proposed CNN-FER includes 32 3x3 filters with a stride of size 1. This hidden layer does includes batch normalization but with no max pooling technique. In the second convolution layer of CNN-FER, has 64 3×3 filters, with the stride of size 1, including batch normalization and max-pooling with a filter size 2×2. The Fully Connected layer has a hidden layer with 512 neurons and Sigmoid activation function as specified in equation (1) owing to the multi labelled classification.

$$f(x) = \begin{cases} \dfrac{1}{1+e^{(-x)}}; & \text{if } x > 0 \\ 0; & \text{if } x \leq 0 \end{cases}$$

(1)

In the proposed CNN-FER system, every image with facial detection having plotted the 68 landmark points is fed as input. The system based on landmark identification, computes the points with $x$ and $y$ coordinates indicated as $(l_x, l_y)$. In general, with 68 landmark points the convolution layer will have $(l_{1x}, l_{1y})$ to $(l_{68x}, l_{68y})$ neurons defined where the landmark $l$ specified as 1 to 68 are the facial landmark points pointed on the detected face. These landmarks are annotated and are labelled from each expression varying in the order of anger, disgust, fear, happy, sadness, neutral, sleep in the implementation. Each point in the landmark is profiled four major vectors for combinatorial analysis such as x-axis, y-axis, angular distance from head central point and angle of landmark [6].

**Emotion prediction through R-CNN:**

The Recurring Neural Network (RNN) differs in the neural network structure by taking a strong advantage of the feedback input from the previous layer. The major advantage of combining CNN with RNN is to solve the problem better when the state of output is static.

This is because of the property of RNN taking activated outputs from the learned classifier to converge into a static solution space in an accurate manner during the training process. The ultimate aim in combining RNN with CNN is to predict the emotion of the children with ASD from expression involving time variant. The emotion of the children is perceived in obtaining continuous analysis of expression over stages. Thus by analyzing the expression over a period of time, the emotion for the next $(t+1)^{th}$ time variant could be predicted and can be classified.

The basic architecture of integrating the CNN with RNN is depicted in Figure 2. On examining the facial expression of the children using CNN with maximized accuracy, the input layer at t-1 of RNN takes the processed expression of the CNN. The RNN network is then feed with input expression evaluated over the time period (t), (t+1), up to (t+n) based on the computational complexity evaluation.

The combined architecture of R-CNN evaluates the facial expression from CNN network, using images captured over the frames. Consequently, this in turn is evaluated by RNN for emotion as text processing system. This combined architecture leads to a better identification of emotion over a specified time period.

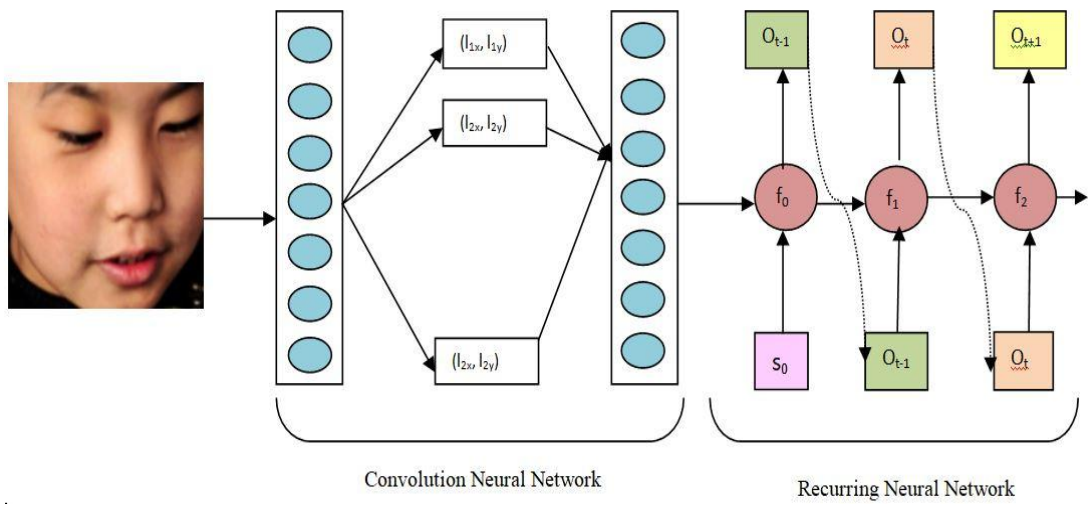

**Figure 2. System Architecture of the integrated CNN and RNN**

The architecture of unfolded recurring neural network sequence explains $x_{t-1}$, $x_t$ and $x_{t+1}$ as the input to the operating function, $S_{t-1}$, $S_t$ and $S_{t+1}$ respectively.

$O_{t-1}$, $O_t$ and $O_{t+1}$ are the output generated from the input activated using $x_i$ and $W$ as weight from the feedback of the previous neuron. $U$ and $V$ are the weight vector for the hidden layer $S_t$ and output layer $O_t$ respectively. The state output of the neuron is computed as in equation (2) and the parameters are mapped to the referred figure.

$$S_t = f(U \times x_t + W \times S_{(t-1)}) \qquad (2)$$

$$Output, \; O_t = Activation \; Function \; (V \times S_t) \qquad (3)$$

The classification based on recurring network is computed as in equation (4) and equation (5). Where, equation (4) specifies the functional evaluation and (5) specifies the output of the current state involving source, function and activated feedback from the previous state.

$$f(t) = g_f(w_i x^{(t)} + w_j f(t-1) + b_f) \qquad (4)$$

$$o(t) = g_o(w_o f(t) + b_{o)}) \qquad (5)$$

where $x$ is the input to the function $f$ at $(t-1)$, $(t)$, $(t+1)$; $f(t)$ is the function that is specified for every layer of RNN say $f_0, f_1, \ldots, f_n$; $w_i$ are the weights acting upon the input $x$ on each layer and $S_0$ is the source weight transferred to the initial input layer of RNN along with computed weight from the previous layer to the next layer specified by time $(t)$ and $(t+1)$; $o(t)$ is the output of each layer at current state that is to be transferred to the next layer and $b$ is the bias value.

The facial expression of the children with and without ASD is examined through a deeper analysis by convolution neural network. Every frame that undergoes the neural network will result in the high prioritized facial expression apparently identified in the child. The expression is saved as the feedback input from time $(t-1)$ to the RNN network that takes the expression of the current frame at time $(t)$. The RNN analyses the expression from the frames over a specified time period to trigger the change in expression possessed by the children. The combination of expression from various frames over the network identifies the emotion perceived by the children.

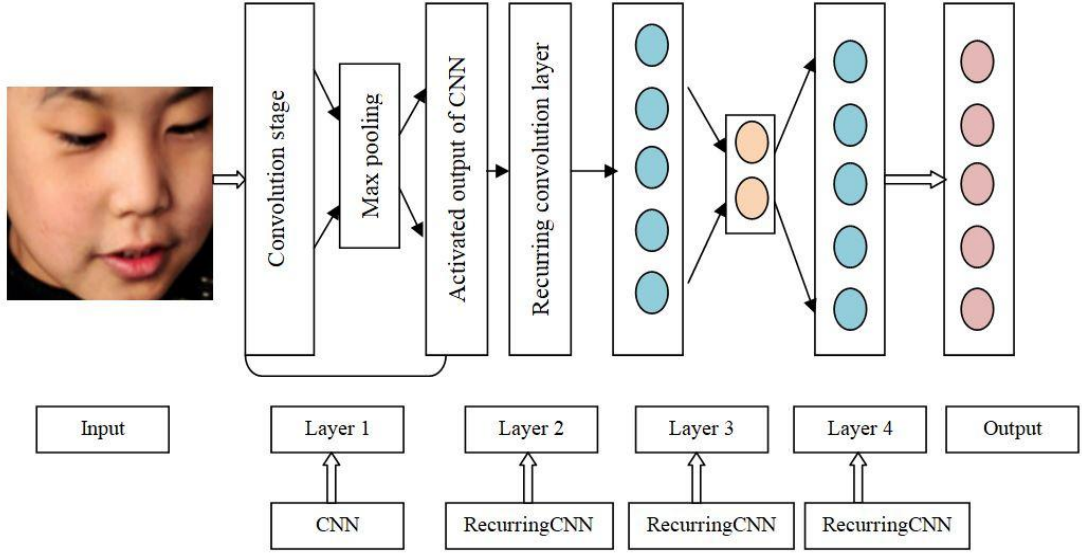

**Figure 3. Working of R-CNN for emotion analysis in children**

The integrated R-CNN for emotional analysis takes the expression with maximized probabilistic ratio for the $(t-1)$ roll as weight of the function to $f(t)$ as specified in equation (4). The $(t^{th})$ layer of the R-CNN computes the expression of the $(t^{th})$ frame and the activating function is guaranteed to operate with the source input of $(t^{th})$ frame and the feedback from $(t-1)^{th}$ frame. The layer 1 of the R-CNN depicted in Figure 3 integrates the CNN with RNN by exhibiting the feedback input commonly termed as weight factor to the next functional computation. The RNN figured out in Figure 3 consists of one hidden layer with max pooling function activated by Softmax function.

$$f(x) = \begin{cases} \dfrac{e^{f_i(x)}}{\sum_{j=1}^{n} e^{f_j(x)}}; & \text{if } x > 0 \\ 0; & \text{if } x \leq 0 \end{cases} \qquad (6)$$

The input to any layered function in R-CNN at time $(t)$ is the expression classified through simple implemented CNN. This is because any frame at $(t-1)^{th}$ time period undergoes a deeper analysis through CNN. The recurring networks pass the maximum probabilistic value from the function towards the next layer of R-CNN. The change in the output expression over the time period $(t)$ is specified to be the emotion in the children and the change of emotion could also be inferred from the intermediate results of the layer. The output of the network at time t is denoted as in equation (7).

$$x_{(t+1)} = \sigma[f(x_t) + g(x_t)] \quad (7)$$

where, $f(x_t)$ and $g(x_t)$ are the weighted functional inputs to the neural network. The output of the function $\sigma$ is denoted as (8)

$$\sigma(x) = \begin{cases} x; & for\ -1 < x < 1 \\ -1; & for\ x \leq -1 \\ 1; & for\ x \geq 1 \end{cases} \quad (8)$$

The proposed system is used to analyse the change in emotion among the children with and without autism. The system also focused to ensure the change within a time interval and the pattern of change in emotion among them.

**DATASET DESCRIPTION**

The data for the analysis of the proposed system in collected from countable children with and without autistic feature and later the sample analysis was also made in analysing videos of autistic children presented over the internet. As the video is generally the sequence of frames, they are processed sequentially. In order to satisfy the processing requirements and the capacity of computation, the dimensions of the input video is chosen to be 320 X 240. Further, the RGB frames are converted to gray scale where a gray scale image consists of only the intensity information of the image rather than the apparent colors. RGB vector is with three dimensional is converted into gray scaled vector is one dimensional with the inculcation of predefined function in OpenCV. After the data acquisition and pre processing based on the computing capacity, the data are fed to the proposed evaluation system. Furthermore, the CNN system as an individual was evaluated over the still images of children with and without autism from Kaggle with a count of 304 and 390 images respectively. The accuracy of the system is proved better over the existing systems [6,7] which adds to the proven improvement for the RCNN FER system.

**Results and Discussions**

The facial expression identified from the real time analyzed ASD children using R-CNN resulted in an accuracy of 91% with an increase of 2% compared to CNN. The average probability of the children with various facial expressions is specified in Table 1 and the maximum perceived expression is found to be disgust, neutral and happy in their respective order. It is also observed that there was no change observed in the maximum perceived facial expression guaranteed from SVM and CNN classifiers rather influencing the classification.

**Table 1. Probability of emotions identified using R-CNN**

| Facial Expressions | ASD Positive (Probability of Expression ) |
|---|---|
| Anger | 0.036045 |
| Disgust | 0.277319 |
| Fear | 0.129893 |
| Happy | 0.144559 |
| Sadness | 0.141052 |
| Neutral | 0.20356 |
| Sleep | 0.070315 |

The emotion of the children with ASD is considered by capturing their facial expression along the time period t, t+10s, t+20s, t+30s respectively. The captured frames for expression analysis are sent through the combined R-CNN that evaluates the change in facial expression to be identified as emotional changes and the pattern of emotion in them. The rate of frames captured could be varied between 1 to n, and the implemented system captures an average amount of frames in a view of not missing any sequence of expression exposed by the children. The frames under the time period are captured continuously at the rate of 30 frames per second. The frames are then dropped in random manner so as to include frames of maximum variations. This also assures minimum computation complexity by reducing the number of frames to be analyzed. Thus by taking frames separated by time period will support for emotional change in the child. Table 2 depicts the probability value of the expression contributing to emotion under a specified time interval.

**Table 2. Probability of various emotions observed during a range of time interval identified using R-CNN**

| Facial expression | At time t | t+10s | t+20s | t+30s |
|---|---|---|---|---|
| Anger | 0.036045 | 0.036154 | 0.046045 | 0.043152 |
| Disgust | 0.277319 | 0.223589 | 0.207319 | 0.18456 |
| Fear | 0.129893 | 0.12785 | 0.129893 | 0.12389 |
| Happy | 0.144559 | 0.16243 | 0.154455 | 0.156711 |
| Sadness | 0.141052 | 0.13984 | 0.151052 | 0.149986 |
| Neutral | 0.20356 | 0.23487 | 0.25356 | 0.25876 |
| Sleep | 0.070315 | 0.07984 | 0.083145 | 0.102457 |

Table 2 portrays the probability of facial expression observed from the autistic child during a specified time interval contributing to emotion pattern. The table also insights that the pattern of expression remains standard over a period of time and the change in the expression towards the time period over the variant is observed to shift from disgust to neutral. To a certain extent, the probability value got maximum biased over sleep and anger in an average. These pattern of emotion could be identified over a periods of time to analysis the maximum emotion perceived by the child and also can be used to determine the pre perceived expression before any stage (especially during aggression). These insights the rate of interest of the child over any specific object or under circumstances.

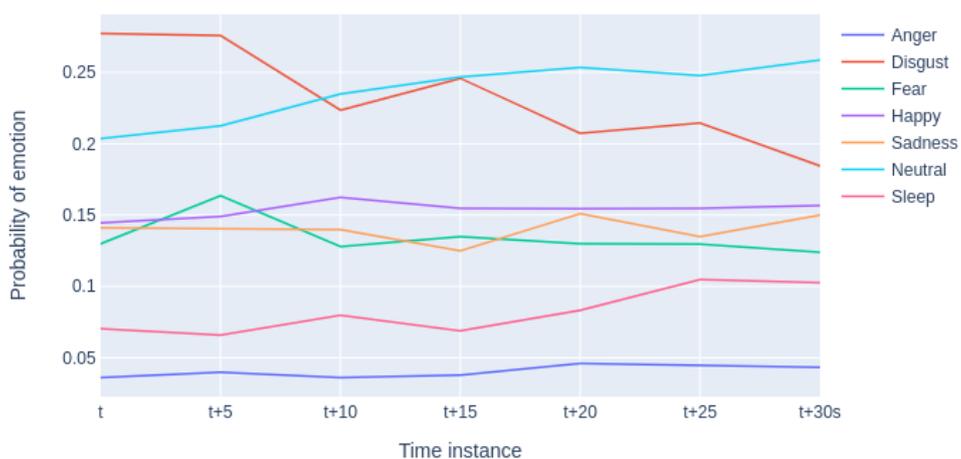

**Figure 4. Probability of emotion observed using R-CNN over a time interval**

Figure 4 depicts the emotion perceived by the children on an average during specified time duration. The figure depicts that there are no sudden fall and rise of emotion in children rather falling into the next highest and nearest emotional probabilistic value. The evaluated countable samples does not show much diverse emotions but there exists various cased that the subject responds vigorous with respect to any disliked objects. Such variations could be analyzed and can be used to control the child when appropriate trigger is made. Every children behavior could be trained under all circumstances so as to ensure better accuracy and reliability of the system.

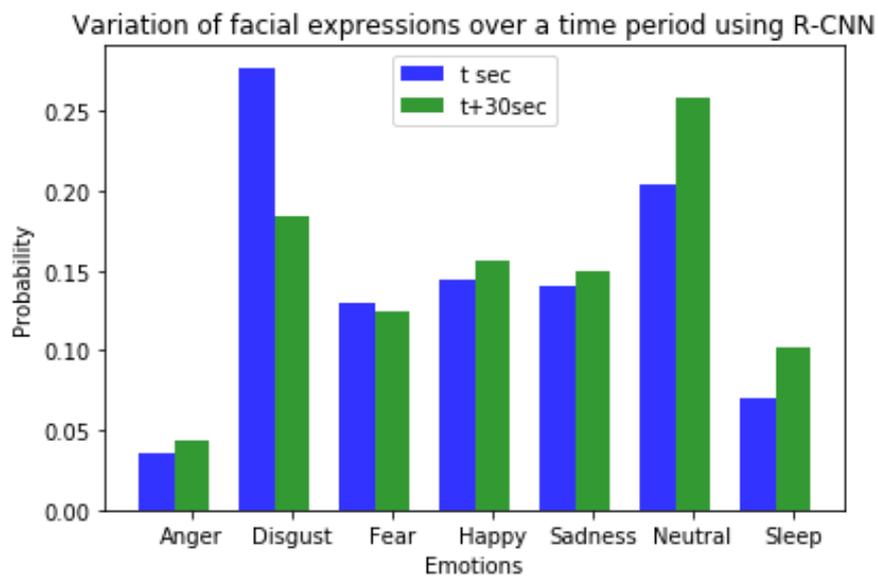

**Figure 5. Rise and fall of emotion observed between the initial and end time variant**

Figure 5 depicts the change in emotion patter observed during time $t$ and $(t+30)\sec$ in autistic child. This insight the change in emotion over a period of time in an average and on further granular analysis patter of emotion in every autistic child could be analyzed to trigger and alert them falling aggressive. This could also be utilized by the parents and caretakers to analyse the duration of change in their child and keep then engaged excluding boredom.

Figure 6 depicts the graphical representation of change of emotion in children with and without autistic characteristics. The values are mapped to the plot on execution of RCNN system and observing the values of emotion for about 50 ticks. The minimum probability emotions could be eliminated and the maximum emotion perceived could be taken into consideration for further clinical analysis by clinicians and occupational therapists.

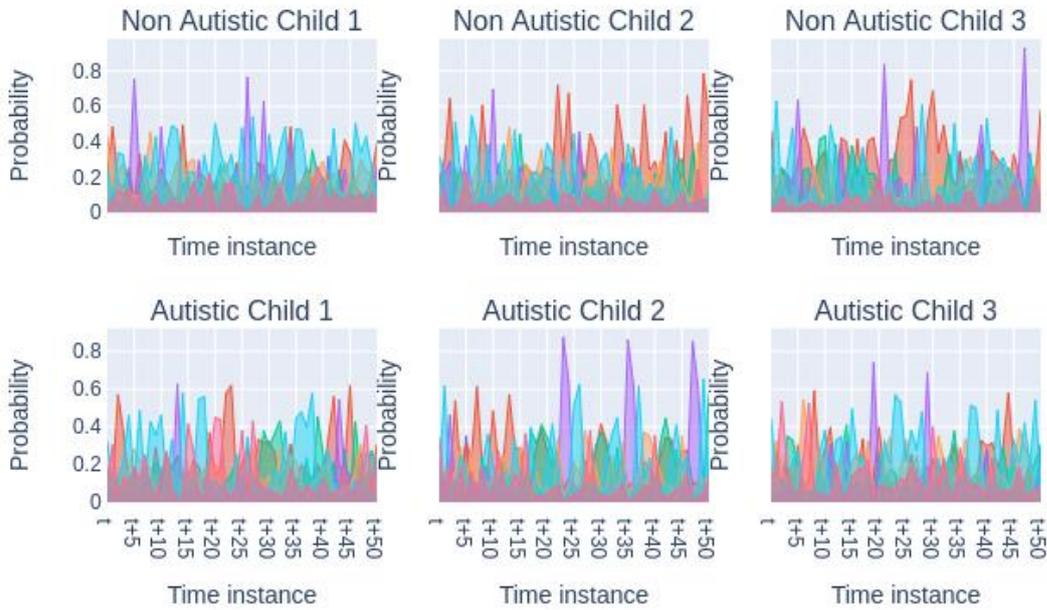

**Figure 6. Probability distribution of emotion in the observed autistic and non autistic children**

**Evaluation of the Model**

On pre processing the acquired data set, the number of images under analysis is reduced. In order to ensure the maximum accuracy of the proposed model, the contribution of k-fold cross validation technique is adopted. However, out of the entire dataset images, the FER system separated 80% of images under training dataset and 20% of the images. The 5 fold cross validation process was incorporated such that, out of the sliced blocks of data, one slice was used for testing the data under every iteration. Similarly, the training data were used to test in either of the epochs. This cross validation method act as a trail method to maximize the training dataset that results in increased accuracy.

Though the number of frames involved in training the ASD positive case stands to be minimum, the dataset is boosted by its volume on applying k-fold method. Also, the exploration of facial expressions stands to be near accurate to the processing capacity. Through the analysis, it is clearly inferred that the majority of the facial expression shown by the either group of children is computed through the summation of prioritized facial expression shown by every individual.

**Accuracy comparison of the proposed models**

The accuracy of the models are evaluated based upon the equation (7)

$$Accuracy = \frac{TP+TN}{TP+TN+FP+FN} \tag{7}$$

The accuracy of emotion analysis in children with ASD is evaluated across the training and test data performance. The emotion analysis undergoes series of iterations in linear prediction from the feedback of the previous layer and the following accuracy was observed as in Figure 7.

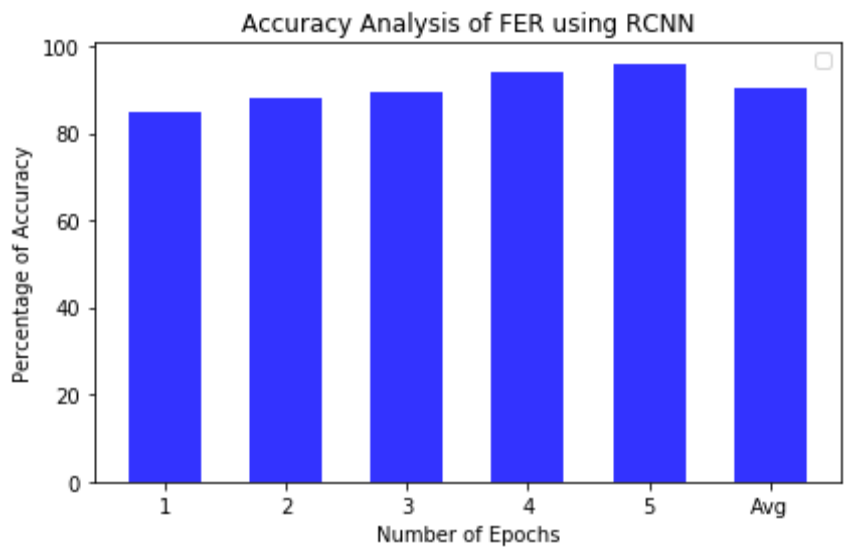

**Figure 7. Accuracy analysis of the proposed classifier over epochs**

On first epoch, the accuracy factor is determined as 0.8462 and in the upcoming iterations of the linear classifier, the accuracy factors were determined as 0.8784, 0.8937, 0.937826 and 0.954157 respectively. On an average, the accuracy of the analysis is found to be approximately 91% in the fifth epoch. The number of iterations is halted owing to the computational complexity involved in executing the neural network. The accuracy could be further improved by increasing the data set size and training the classifier with much more varied and deeper feature combinations.

The following Figure 8 represents the accuracy comparison of the proposed models for FER systems namely SVM, CNN and R-CNN at a specified time t. Though there is no change in the probability of the facial expression compared over the models, there is a strong advent to the influence of the facial expression identified within the ASD +ve and ASD -ve children. These models could be used as an early biomarker of autism identification and exploration of facial expression in them.

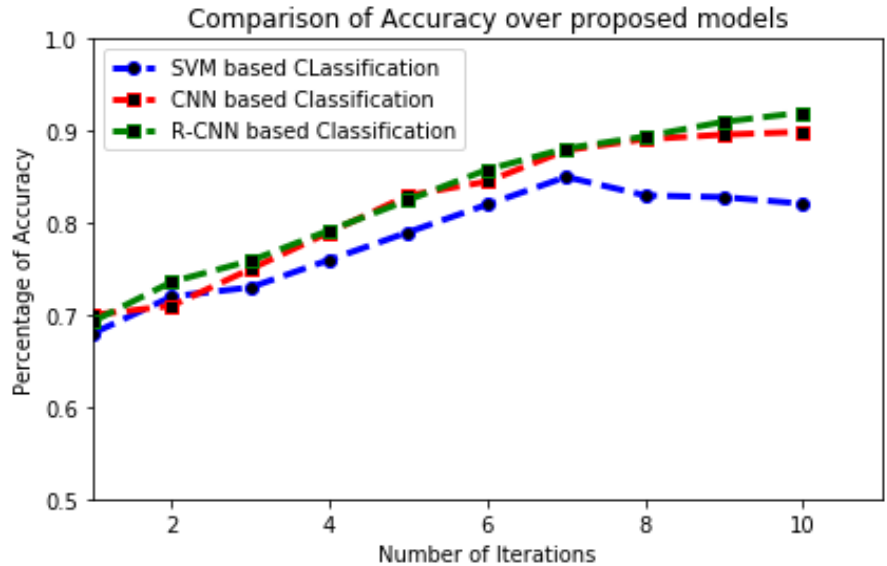

**Figure 8. Comparison of Accuracy on R-CNN over existing systems**

**Conclusion and Future Scope**

The integrated R-CNN based FEA system identifies the emotion of the children involving time variant using live video input. The system proves better improvement in expression analysis and also suggests the point of time when there exists a change in expression. The system indicates that the maximum expression perceived by the autistic children is disgust, neutral and anger. The fall of expression over time for the autistic children is also into the perceived expression which supports for the identification of repetitive behaviour in the autistic children. The system proves better in accuracy of 91% when compared to the other existing models for facial expression evaluation. Also, the major contribution of this proposed system is to identify the autistic character in the child at an early stage by direct evaluation of the child behaviour rather through questionnaires. This pattern of evaluated expression incurred in every child could also be inferred supporting early autism identification. The future scope of the paper lines in altering the family members in prior to the emotional reaction of the children with proper accompanied IoT systems. This real time intelligent surveillance can improve in the development stages of the child before falling into aggression and hatred state.